\documentclass{article}

\usepackage{microtype}
\usepackage{graphicx}
\usepackage{booktabs} 

\usepackage{hyperref}

\usepackage[preprint]{icml2026}

\usepackage{amsmath}
\usepackage{amssymb}
\usepackage{mathtools}
\usepackage{amsthm}

\usepackage[capitalize,noabbrev]{cleveref}

\usepackage[most]{tcolorbox}
\usepackage{enumitem}
\usepackage{xcolor}
\usepackage{subfigure}
\usepackage{tabularx}
\usepackage{makecell}

\def\subsectioncrefname{Section}

\newcommand{\ourmethod}{CoScale-RL}
\newcommand{\mytitle}{\ourmethod: Efficient Post-Training by Co-Scaling Data and Computation}

\theoremstyle{plain}
\newtheorem{theorem}{Theorem}[section]

\newtheorem{lemma}[theorem]{Lemma}

\theoremstyle{definition}

\theoremstyle{remark}

\usepackage[disable,textsize=tiny]{todonotes}

\icmltitlerunning{\mytitle}

\begin{document}

\twocolumn[
  \icmltitle{\mytitle}



  \icmlsetsymbol{equal}{*}

  \begin{icmlauthorlist}
    \icmlauthor{Yutong Chen}{thu}
    \icmlauthor{Jiandong Gao}{thu}
    \icmlauthor{Ji Wu}{thu,cai,bt}
  \end{icmlauthorlist}

  \icmlaffiliation{thu}{Department of Electronic Engineering, Tsinghua University, Beijing, China}
  \icmlaffiliation{cai}{College of AI, Tsinghua University, Beijing, China}
  \icmlaffiliation{bt}{Beijing National Research Center for Information Science and Technology, Beijing, China}
  \icmlcorrespondingauthor{Jiandong Gao}{jdgao@tsinghua.edu.cn}
  \icmlcorrespondingauthor{Ji Wu}{wuji\_ee@tsinghua.edu.cn}

  \icmlkeywords{Reinforcement Learning, Large Language Model, Post-Training, Scaling Law, Data Efficiency}

  \vskip 0.3in
]



\printAffiliationsAndNotice{}  

\begin{abstract}

Training Large Reasoning Model (LRM) is usually unstable and unpredictable, especially on hard problems or weak foundation models. We found that the current post-training scaling strategy can still improve on these cases.
We propose \ourmethod, a novel scaling strategy with better data and computational efficiency. 
We first scale up solutions to make problems solvable. The core idea is to collect multiple solutions for each problem, rather than simply enlarging the dataset. Then, we scale up rollout computation to stabilize Reinforcement Learning.
We further leverage a model merge technique called Re-distillation to sustain or even improve computational efficiency when scaling up. 
Our method significantly improves data and computational efficiency, with an average 3.76\(\times\) accuracy improvement on four benchmarks. \ourmethod\ is able to improve an LRM's ability boundary without an extensive SFT dataset.
Our method provides a new scaling direction to further improve LRM's reasoning ability.

\end{abstract}

\section{Introduction} \label{introduction}

Large Reasoning Model (LRM) has shown remarkable reasoning ability by leveraging post-training, which often involves Supervised Fine-Tuning (SFT) and Reinforcement Learning (RL) \citep{guo_deepseek-r1_2025,team_qwq-32b_2025,kimi_team_kimi_2025}. However, the current post-training paradigm encounters two critical challenges. The first is on improving LRM's ability boundary. Recent research has shown that LRM can only learn those problems within the fundamental ability boundary \citep{yue_does_2025,zhao_echo_2025}. A common approach is combining SFT and RL \citep{yan_learning_2025,nath_adaptive_2025,qin_supervised_2025}. While SFT or the distillation technique is found useful for small LRM \citep{face_open_2025}, it is common that finetuning does not improve performance if the dataset is small, or even hurts the RL potential \citep{chu_sft_2025,chen_synergy_2025,yu_reassessing_2025,kang_quagmires_2025}. 

The second challenge is the instability of RL on weak models or difficult reasoning tasks. Studies found that RL on hard problems often failed, or became unstable after training more steps \citep{gao_soft_2025,jin_search-r1_2025,deng_grpo_2025}. Recent works have found some reasons, such as all-zero rewards \citep{yu_dapo_2025}, training-inference gap \citep{zheng_stabilizing_2025}, and dynamic difficulty adjustment \citep{khatri_art_2025}. However, RL instability remains an open problem, especially on weak Language Models or too hard problems. 

\begin{figure}[t]
  \centering
  \includegraphics[width=0.48\textwidth]{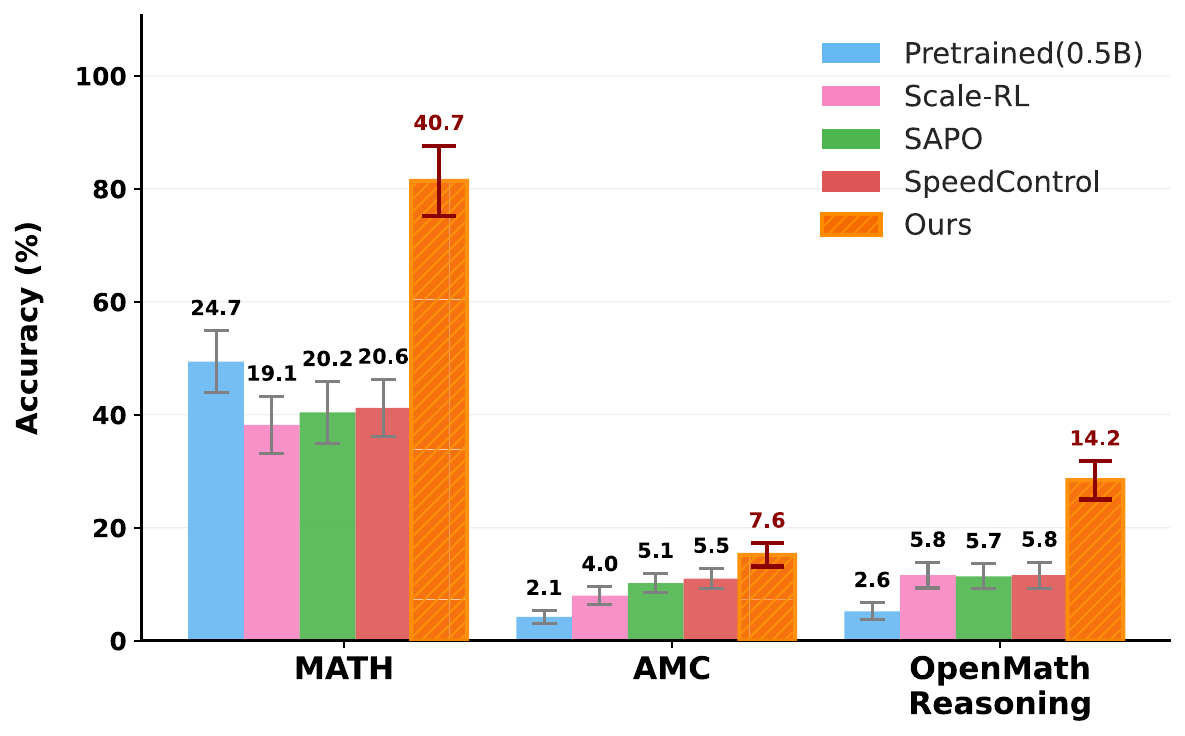}
  \caption{\ourmethod\ consistently outperforms baselines on multiple benchmarks. Compared with the Pretrained (0.5B), our method has a 3.76\(\times\) average improvement on four math reasoning benchmarks. Refer to \cref{sec:exp-large-scale} for details. Error Bar: 95\% CI.}
  \label{fig:large-scale}
\end{figure}

In this paper, we focus on a typical post-training setting with both SFT and RL: The SFT dataset contains multiple problem-solution pairs. Each \textit{problem} has a verifiable answer. Each \textit{solution} contains a long reasoning trace with a predicted answer. The solution is usually generated by a human or a stronger LRM. In SFT, the model is finetuned on solutions with corresponding problems as input. In RL, there is a \textit{rollout} phase followed by a \textit{policy update} phase. The model will generate \(N\) solutions for each problem in the rollout phase. We call the number of generated solutions for each problem as \textit{Rollout N}. The problem set of RL can be different from SFT. Each generated solution will be assigned a \textit{reward} based on the correctness. In the policy update phase, the model is trained on generated solutions. 

We found that there is a basic but important assumption that is often overlooked in recent studies: \textbf{LRM may not learn correctly with just one solution}. 
This is a crucial factor in post-training, because learning from a single solution can not guarantee a high enough accuracy, which makes RL impractical. For example, if a model has 0.1\% accuracy on a hard problem, it may need Rollout N$\geq$10,000 to obtain a stable gradient, which is hardly reached in real cases. When one problem has such an instability, it may damage the inner structure of the model and even cause an RL failure.

\begin{figure}[t]
  \centering
  \includegraphics[width=0.40\textwidth]{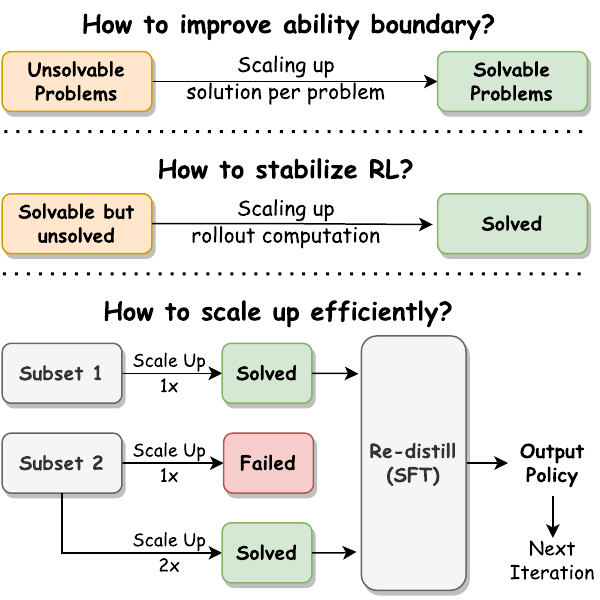}
  \caption{Three core techniques in \textbf{\ourmethod}. The definition of Solvable/Unsolvable/Solved and other details can be found in \cref{sec:method}.}
  \label{fig:core-ideas}
\end{figure}

Our main contribution is to propose a new scaling direction. We found that \textbf{simultaneously scaling up \textit{Solution per Problem} and \textit{Rollout N}} (i.e., the number of rollout solutions for each problem) is an effective way to solve the above issues. As described in \cref{fig:core-ideas}, we collect fewer problems but with more solutions for each problem. We found that SFT on multiple solutions is surprisingly effective in improving reasoning performance, even with the same dataset size. Then, we scale up Rollout \(N\) to further reduce noise. By co-scaling data and computation, our method is able to directly train on even the hardest problems.

One of the potential issues of our scaling strategy could be reducing the overall efficiency, because some easy problems do not need many resources. We address this by assigning different scaling coefficients for different problems. Specifically, we divide all problems into multiple groups. Each group will run RL separately with different hyperparameters, but will be combined after training. To achieve this, we introduce Re-distillation \citep{chen_towards_2025} as a model merge method, which efficiently merges all RL progresses with far less computation. While the idea of distillation or model merging is investigated in many recent works \citep{wilhelm_distilling_2025,wu_unlocking_2025}, we further apply this technique in improving RL efficiency. 

Our contributions are listed as follows:

\begin{itemize}
  \item We find that scaling up the solution per problem significantly improves accuracy on the training set compared with simply enlarging the dataset size. The advantage of SFT on multiple solutions is also generalizable to other multiple math benchmarks (\cref{sec:exp-large-scale}).
  \item We propose \ourmethod, a better scaling strategy for post-training. Our method has higher data and computational efficiency compared with general RL settings (\cref{sec:exp-data-efficiency} and \cref{sec:exp-compute-efficiency}). With \ourmethod, we successfully break the ability boundary for a 0.5B Language Model on long reasoning problems with about 10K tokens per response (\cref{sec:Q1}).
  \item We theoretically analyzed how accuracy and other critical hyperparameters influence RL computational efficiency (\cref{sec:rl-scaling}). Our method provides a practical solution to address RL instability by Data \& Algorithm co-design. 
\end{itemize}

\section{Related Works}

\subsection{RL Ability Boundary}

Studies found that Reinforcement Learning on LRMs often failed to improve or even shrink the model's ability boundary. \citet{yue_does_2025} found that while SFT models have weak Pass@1, they outperform RL models on Pass@K, suggesting RL narrows the ability boundary. \citet{zhao_echo_2025} similarly observed that RL sharpens the output distribution rather than broadening reasoning. Several methods aim to mitigate this problem. \citet{chen_passk_2025} and \citet{walder_passk_2025} directly optimize for Pass@K. Some works improve Pass@K by enhancing exploration or deriving augmented problems \citep{liang_beyond_2025,hu_brorl_2025}. Other approaches leverage SFT to broaden potential when applying RL \citep{qin_supervised_2025, nath_adaptive_2025, yan_learning_2025, li_limr_2025}. Additionally, Pro-RL \citep{liu_prorl_2025, hu_prorl_2025} claims prolonged RL training can itself improve the ability boundary. Overall, it is still challenging to improve the ability boundary without an extensive SFT dataset.

\begin{figure*}[ht]
  \centering
  \includegraphics[width=0.99\textwidth]{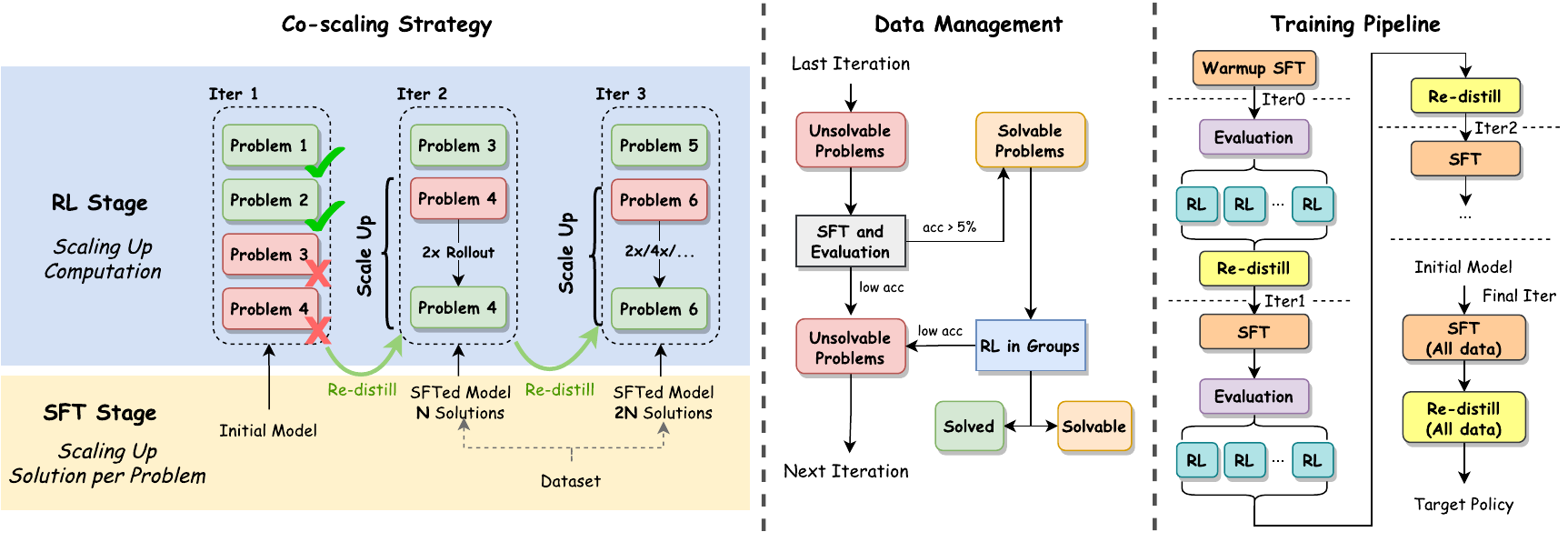}
  \caption{Overview of our method. \textbf{Co-scaling Strategy (Left)}: We simultaneously scale up \textit{Solution per Problem} in SFT and \textit{Rollout N} in RL. After entering a new iteration, we gather more solutions for unsolvable problems. Once a problem becomes solvable, we scale up computation until RL solves this problem. \textbf{Data Management (Middle)}: All problems are labeled as unsolvable at start. After evaluation, we label those problems as solvable with enough accuracy. We only collect solvable problems for RL. \textbf{Training Pipeline (Right)}: We finetune the initial model with a few problems to switch output mode. Then we follow the same iterative process. In the final iteration, we retrain the model from the initial model to avoid catastrophic forgetting.}
  \label{fig:pipeline}
\end{figure*}

\subsection{Scaling Law of Reinforcement Learning}

The scaling law for RL differs from pretraining. \citet{khatri_art_2025} found RL scales differently, and final accuracy can be further predicted from partial reward curves. RL Scaling can be applied in both training and inference. \citet{cen2025webscalerlautomateddatapipeline} converts pretrain data into question-answer pairs for RL. Training-free methods scale up response length during inference. \citet{muennighoff_s1_2025} appends `Wait' tokens to prolong inference. \citet{chalumeau_breaking_nodate} identified inference scaling as the key to breaking ability boundaries by using extra computation or leveraging a multi-agent system.

\subsection{Computational Efficiency}

Improving computation efficiency is related to the complex training dynamics of the Language Model. The Stochastic Differential Equation (SDE) is a powerful tool to analyze the training dynamics. \citet{li_stochastic_2019} made a theoretical foundation of SDE application on common scenarios. \citet{compagnoni_adaptive_2025} generalized SDE to adaptive algorithms such as Adam. \citet{malladi_sdes_2022} theoretically proved that the training process can be accelerated or slowed down by adjusting hyperparameters, especially the batch size and learning rate.

\section{Methods} \label{sec:method}

We describe how \ourmethod\ works in \cref{sec:co-scale}. Our method includes the co-scaling strategy, data management, and the training pipeline. In \cref{sec:rl-scaling}, we explain why we choose to scale up Rollout \(N\) through a theoretical perspective. Before introducing our method, we first define some important concepts here: 

\textit{Data Scaling}: In our method, we use data scaling to indicate increasing the number of solutions per problem in the SFT stage.

\textit{Computational Scaling}: In our method, we use this term to indicate increasing \textit{Rollout N} in the RL stage, which needs more generations in each step.

\textit{Solvable / Unsolvable / Solved}: In our experiments, we label a problem as \textit{Solvable} if its accuracy is between 5\% and 70\%. It is \textit{Unsolvable} if the accuracy is below 5\%, or \textit{Solved} if over 70\%. 

\subsection{\ourmethod : Co-scaling Data and Computation} \label{sec:co-scale}

We propose \ourmethod\ to efficiently scale up data and computation. As described in \cref{fig:pipeline} (Left), \ourmethod\ is an iterative algorithm. There is an SFT stage and a following RL stage in each iteration. In the SFT stage, we add new solutions for all unsolvable problems to scale up the solution per problem. After SFT, we evaluate each unsolved problem's status by sampling some solutions and estimating the accuracy. Only solvable problems could enter the RL stage. In the RL stage, we split problems into multiple groups and scale up Rollout \(N\) when any group fails. Each group is trained independently.

\textbf{Data Management}: \ourmethod\ maintains four problem pools: Unsolvable, Solvable, Solved, and Failed. Initially, all problems are in the Unsolvable set. We SFT with more solutions for unsolvable problems and compute accuracy by generating solutions. Those problems with enough accuracy (e.g., above 5\%) will enter the Solvable set. Only solvable problems will be used for RL training. After RL, those problems that achieved high enough accuracy (e.g., above 70\%) will enter the Solved set. Other problems that are still in progress will return to the Solvable set. Those who fail in the current RL stage will be put into the Failed set if they reach the budget limitation, which means they will no longer enter the following iterations. For those problems that failed but still have a budget, they will return to Unsolvable set. At last, all problems will be either solved or failed. Refer to \cref{appendix:alg-desc} for a formal algorithm description.

\textbf{Model Merge}: We leverage Re-distillation \cite{chen_towards_2025} to address several technical issues in \ourmethod. The primary function of Re-distillation is to compress the performance gain from RL into an SFT dataset. Specifically, for each problem, we collect the last 100 correct solutions from the RL replay buffer and fine-tune the initial policy. By converting RL into SFT, it is possible to combine multiple independent RL processes once they share the same initial policy. 

We found that Re-distillation inherently achieved failure resistance and dynamic hyperparameter assignment. By dividing RL into multiple independent processes, it is possible to restart any failed process without affecting others. It also enables assigning different hyperparameters to different subsets, which is critical to maintain computational efficiency.

\textbf{Training Pipeline}: The concrete training pipeline is depicted in \cref{fig:pipeline} (Right). We start from a warm-up SFT, which converts the output mode into deep thinking mode. In each iteration, we finetune the model based on previous progress but do not restart from scratch. While this removes redundant computation, it also introduces a new issue called \textit{Catastrophic Forgetting}. The policy may forget how to solve problems in earlier iterations. To address this issue, in the final iteration, we retrain the policy from the initial model. 

\subsection{Theoretical Analysis for Efficient Computational Scaling} \label{sec:rl-scaling}

We theoretically proved that \textbf{the computational efficiency can be described as a quadratic-equation of \(\eta/N\) (Learning Rate / Rollout N)}. This explains why we choose to change Rollout \(N\) but not other hyperparameters. 
As a start, we define the computational efficiency in \cref{equ:comp-eff-def}, which measures the contribution of each rollout solution for the performance improvement.

\begin{equation} \label{equ:comp-eff-def}
  \text{Computation Efficiency } \mathcal{E} := \frac{\Delta \text{Accuracy}}{N_{\text{Rollout}}}
\end{equation}

Inspired by previous works \citep{li_surge_2024,li_stochastic_2019,mccandlish_empirical_2018}, we consider three basic factors which may be critical: learning rate \(\eta\), rollout \(N\), and correct probability \(p\) of each problem. We want to describe how these three factors influence computational efficiency.

\subsubsection{Optimal \(\eta/N\) to Maximize Efficiency}

Our first goal is to maximize \(\mathcal{E}\) by adjusting \(\eta\) and \(N\). Since the training dynamic is highly non-linear, we predict the linear effect for each step, which can be described by Stochastic Differential Equation (SDE). When the learning rate \(\eta\) is small enough, SDE can be an effective approximation of real-world training dynamics. In \cref{theorem1}, we proved that computational efficiency is a quadratic-equation of \(\eta/N\). See \cref{sec:appendix-proof-comp-optimal} for the complete proof. 

\begin{theorem} \label{theorem1}
  Under some assumptions, the computational efficiency of both SGD and Adam Optimizer can be described as: 
  \begin{center}
    \(\mathcal{E} = \frac{\eta}{N}\mu_p -  \left ( \frac{\eta}{N} \right )^2 \mu_n \)
  \end{center}
  where \(\mu_p\) and \(\mu_n\) are not related to \(\eta\) or \(N\).
\end{theorem}

When \(\eta/N\) corresponds to the minimum point, we obtain optimal efficiency. When \(\eta/N\) is sufficiently small, the computational efficiency can always be positive, which means RL can be stabilized. Using a too large learning rate or too small \(N\) will lead to negative \(\mathcal{E}\), which may make RL unstable.

\subsubsection{The Effect of Problem Difficulty}

Here we further consider accuracy \(p\) in training dynamics. A common phenomenon is that RL usually failed on overly difficult problems. One possible reason may be a too small Rollout \(N\). Intuitively, when \(p\) is very small and Rollout \(N\) is also small. Most rollout samples will become invalid because they have all zero rewards. When \(N\) is large enough, it is easier to find one positive solution. Here we introduce \(\hat N(p)\) to evaluate how actual \(N\) will be affected by problem difficulty.

\begin{equation} \label{equ:prob-effect}
  \hat N(p) = N \;\mathbb{P}[\text{Var}(R)\neq 0] = N[1-p^N-(1-p)^N]
\end{equation}

In \cref{equ:prob-effect}, when we generate \(N\) solutions for a problem with probability \(p\) to be correct, we actually only generate \(\hat N\) solutions with non-zero gradient on average. Combined with \cref{theorem1}, the real \(\eta/\hat N(p)\) will become larger in policy update. A simple but effective way is to increase \(N\) and decrease \(\eta\). When \(N\) is large enough, \(\hat N(p)\) will converge to \(N\) rapidly, which keeps computational efficiency stable during training.

\subsubsection{Takeaways to Maximize Computational Efficiency} \label{sec:takeaways}

Overall, the following guidance tells how to improve computational efficiency theoretically. Here, we ignore practical challenges. We discuss how our method leverages these theoretical results practically in \cref{sec:practical-alg}.

\textbf{Remove unsolvable problems in the RL dataset}. Training on all-zero rewards does not give any valid gradient but wastes resources. 

\textbf{Keep large Rollout \(N\) for difficult problems}. When \(N\) is large enough, \(\hat N(p)\) will converge to \(N\) rapidly, which stabilizes the overall efficiency. Otherwise, RL training may fail even with smaller \(\eta\) or more steps.

\textbf{Decrease \(\eta/N\) to stabilize RL training}. A larger \(N\) or smaller \(\eta\) will always stabilize RL training, but too small \(\eta/N\) will reduce computational efficiency.

\textbf{Keep \(\eta/N\) close to the optimal point}. Theoretically, each problem has its own optimal \(\eta/N\), which may dynamically change during the RL process. It is recommended to decrease \(\eta/N\) when RL fails or gets stuck.

\begin{table*}[ht]
\centering
\begin{small}
\caption{Performance Comparison on Math and General Reasoning Benchmarks (\% \(\pm\) Standard Deviation).}
\label{tab:main-results}
\setlength{\tabcolsep}{3pt}
\begin{tabular}{lccccc}
\toprule
                & \multicolumn{4}{c}{\textbf{Math Reasoning} (Selected Combinatorial and Probabilistic Problems)} & \textbf{General Reasoning} \\
\cmidrule(lr){2-5}
\cmidrule(lr){6-6}
\textbf{Model}  & \textbf{MATH-500} & \textbf{AMC} & \textbf{OpenMathReasoning} & \textbf{Olympiad Math}    & \textbf{Reasoning GYM} \\
\midrule
\multicolumn{6}{c}{\textbf{Main Results (\%)}} \\
\midrule
Qwen2.5-0.5B Instruct   & $24.7 \pm 1.4$    & $2.1 \pm 0.29$ & $2.6 \pm 0.39$ & $0.29 \pm 0.09$ & $6.0 \pm 1.1$ \\
Pretrain + RL           & $20.8 \pm 0.5$   & $2.7 \pm 0.33$ &  $3.5 \pm 0.46$ & $0.45 \pm 0.11$ & $8.6 \pm 1.3$ \\
SFTed (w/ Data Scaling) & $23.2 \pm 1.4$    & $1.4 \pm 0.24$ & $4.3 \pm 0.51$ & $0.24 \pm 0.08$ & $5.4 \pm 1.1$ \\
\textbf{Ours (\ourmethod)}  & $\mathbf{40.7 \pm 1.6}$ & $\mathbf{7.6 \pm 0.53}$ & $\mathbf{14.2 \pm 0.87}$ & $\mathbf{1.26 \pm 0.18}$ & $6.8 \pm 1.1$ \\
\midrule
\multicolumn{6}{c}{\textbf{Baselines (\%)}} \\
\midrule
SFTed
(w/o Data Scaling)                   & $17.9 \pm 1.3$  & $2.3 \pm 0.30$  & $2.5 \pm 0.40$  & $0.51\pm0.17$   & $5.8 \pm 1.0 $ \\
GRPO 
\citep{shao_deepseekmath_2024}       & $16.7 \pm 1.3$ & $5.1 \pm 0.44$  & $5.6 \pm 0.57$  & $0.46\pm0.12$ & $8.4 \pm 1.2$ \\
Scale-RL 
\citep{khatri_art_2025}              & $19.1 \pm 1.3$ & $4.0 \pm 0.40$ & $5.8 \pm 0.58$ & $0.29 \pm 0.08$ & $5.4 \pm 1.0$ \\
COMPASS 
\citep{chalumeau_combinatorial_2024} & $17.0 \pm 1.2$ & $6.8 \pm 0.51$  & $5.4 \pm 0.57$  & $0.72 \pm 0.13$ & $6.8 \pm 1.1$ \\
SAPO 
\citep{gao_soft_2025}                & $20.2 \pm 1.4$ & $5.1 \pm 0.44$ & $5.7 \pm 0.57$ & $0.48 \pm 0.11$ & $\mathbf{8.8 \pm 1.2}$ \\
SpeedControl 
\citep{lin_controlling_2025}         & $20.6 \pm 1.3$ & $5.5 \pm 0.46$ & $5.8 \pm 0.59$ & $0.45 \pm 0.11$ & $7.4 \pm 1.17$ \\
\bottomrule
\end{tabular}
\end{small}
\end{table*}

\section{Experiments}

We conducted a large-scale experiment to show that \ourmethod\ can improve performance on multiple benchmarks. We choose training on combinatorial and probabilistic problems, a subfield of math that requires complex reasoning ability. The training set contains 4.7K problems and 30K traces in total. During inference, each output sequence is about 8K tokens. 

We evaluate our method on four math benchmarks (MATH \citep{hendrycks_measuring_2021}, AMC12 (from \citet{numinamath}), OpenMathReasoning \citep{moshkov_aimo-2_2025}, OlympiadMATH \citep{sun_challenging_2025}) and one general reasoning benchmark, Reasoning GYM \citep{stojanovski_reasoning_2025}. Four math benchmarks are filtered to keep combinatorial and probabilistic problems only.

Different from existing works, we choose Qwen2.5-0.5B Instruct \citep{qwen_qwen25_2025}, an extremely weak model without math-specific finetuning, as the base model. Starting from such a Language Model is often more challenging than an LRM or one finetuned with specialized knowledge, such as R1-distilled models \citep{guo_deepseek-r1_2025} or Qwen2.5-MATH \citep{qwen25math}.

\subsection{Experimental Settings} \label{sec:exp-settings}

\ourmethod\ uses GRPO \citep{shao_deepseekmath_2024} with adaptations described in \cref{sec:rl-importance-sampling}. We listed the general settings as follows. See \cref{sec:appendix-exp-details} for details.

\textbf{Training Dataset}: We choose OpenMathReasoning and filter a subset for training . First, we prompt Qwen2.5-32B Instruct to collect combinatorial and probabilistic problems. Then we filter out solutions that are not verifiable (i.e., proofs) or too long. There are 5K valid problems left after filtering. These problems are randomly split into 3 subsets: \textit{OMR-seed} (90 problems), \textit{OMR-rest} (4710 problems), and \textit{OMR-test} (200 problems). When scaling up solutions per problem, we first use existing solutions, then prompt QwQ-32B \citep{team_qwq-32b_2025} to generate new solutions. 

\textbf{Test Datasets}: For the test set, we filter all combinatorial and probabilistic problems in MATH-500, AMC, and OlymMATH. Problems in Reasoning GYM are synthesized by default settings. The test set includes MATH-500 (54 problems), AMC12 (19 problems), \textit{OMR-test} (200 problems), OlymMATH (29 problems), and Reasoning GYM (500 problems, no filtering). These test sets are duplicated up to 128 times to reduce estimation variance. 

\textbf{Reward Design}: We use a binary reward in all experiments. To judge correctness, we prompt Qwen2.5-32B Instruct with questions, ground truth answer, and extracted answer from the last boxed\{\}. We use greedy decoding in reward evaluation.

\textbf{Evaluation Metric}: We compute the average Pass@1 rate on each test set. The decoding setting is temperature 0.7, top-p 0.95, and max token length 25K. There is no extra information for test-time scaling. Refer to \cref{sec:appendix-eval-details} for evaluation details.

\subsection{Baseline Settings} \label{sec:baseline-settings}

We implemented four baseline methods, including RL scaling strategy, RL algorithm, inference sampling strategy, and test-time reasoning enhancement. All baseline methods are based on the SFTed model without our data scaling strategy. See \cref{sec:appendix-baseline-details} for details.

\textbf{SFTed Model} (w/o Data Scaling): We use the same amount of data and settings to train an SFTed model without our data scaling method. To create a dataset, we use all problems in \textit{OMR-rest} dataset and add new problems from OpenMathReasoning. Each problem has only one correct solution. The total problem set is 32K, which equals what we used in training \ourmethod. 

\textbf{Scale-RL} \citep{khatri_art_2025}: We follow the best practice of Scale-RL, which includes CISPO \citep{cispo} as the loss function, advantage batch norm, FP32 LM head, zero std sample selection, and curriculum learning. 

\textbf{SAPO} \citep{gao_soft_2025} and \textbf{GRPO}: We implement SAPO algorithm with \(\tau_{\text{pos}}=1.0\) and \(\tau_{\text{neg}}=1.05\). The GRPO algorithm is implemented in the same way as what we used in \ourmethod.

\textbf{COMPASS} \citep{chalumeau_combinatorial_2024}: We create compass vectors with 16 dimensions between -1 and 1. In the training phase, we select the best compass vector from 8 candidates. In the inference phase, we allow the model search 500 times for each problem. For fairness, the final result is randomly sampled based on the last CMA vector.

\textbf{SpeedControl} \citep{lin_controlling_2025}: We extracted the speed controlling vector from the best baseline (SAPO). Then we search coefficient \(\alpha\) for the best accuracy improvement. This method is inference only. 

\subsection{Results} \label{sec:exp-large-scale}

We train Qwen2.5-0.5B Instruct by our proposed pipeline. For data scaling, we scale up the solution per problem by 0, 2, and 8. For computational scaling, we manually scale up Rollout \(N\) if the training reward is on a plateau. Limited by our computation budget, the training stopped after 3 iterations. We list the main results in \cref{tab:main-results} and \cref{fig:large-scale}.

\textbf{\ourmethod\ significantly improves ability boundary}:
Our method outperforms other baselines not only on OpenMathReasoning (14.2\% vs 5.8\%), but also on other out-of-distribution benchmarks like MATH-500 (40.7\% vs 24.7\%) and AMC12 (7.6\% vs 6.8\%). We found that the improvement of \ourmethod\ can not be reproduced by simpler methods such as RL on Qwen2.5-0.5B Instruct, or directly SFT without RL. Simply scaling either direction, such as computation (Pretrain+RL) or data (SFTed w/ data scaling), does not improve performance compared with the pretrained model. Scaling in both directions released the potential. For example, \ourmethod\ reached 40.7\% on MATH-500, which can not be achieved by other methods. However, we also found that such improvement is field-related. On general reasoning tasks (Reasoning GYM), \ourmethod\ does not behave as superior as combinatorial and probabilistic problems.

\textbf{Co-Scaling gets more bonus in RL stage}:
We also compare our method with other scaling strategies. With equal dataset size, our data scaling method does not immediately improve performance after SFT, but benefits more in the RL stage. The reason that SFT does not improve performance is probably that the reasoning length is significantly longer than the base model's ability, which prevents performance transfer. However, the surprising improvement in RL indicates that scaling up the solution per problem is an effective way to improve RL potential. This is probably because our method improves the chance of solving more challenging problems due to our data scaling strategy (See \cref{sec:exp-data-efficiency} for comparison).

\textbf{Scaling is more important than algorithmic design}:
Comparing different baseline RL algorithms (SAPO, Scale-RL, GRPO), we found that there is no significant difference in overall performance. The influence of scaling outweighs algorithmic factors as well as other test-time scaling methods such as COMPASS and SpeedControl. This is probably because Qwen2.5-0.5B Instruct is a well-trained Language Model with sufficient post-training on short responses. Therefore, it is challenging to further improve the performance without introducing extra SFT data.

\section{Ablation Studies}

We made four ablation experiments to investigate how each design of \ourmethod\ contributes to the final performance. Refer to \cref{sec:appendix-ablation} for details of ablation settings.

\subsection{Can LLM Solve Hard Problems with a Few SFT Solutions?} \label{sec:Q1}

The most important assumption in our paper is that LLM can solve challenging problems with a few solutions, even if it is out of its ability boundary. We made a tiny experiment to prove that learning an `out-of-boundary' problem does not require an extensive dataset. 

\begin{figure}[htb]
  \centering
  \includegraphics[width=0.45\textwidth]{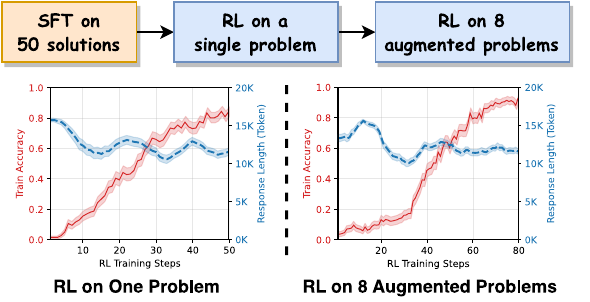}
  \caption{\textbf{1.5B LLM can solve an AIME level problem after being finetuned on 50 solutions.} \textbf{Upper}: Training pipeline. \textbf{Lower} (Left): RL on only one problem. We found a stable improvement in reward and a steady 10k level response length. \textbf{Lower} (Right): RL on 8 augmented problems. RLed model succeeded in solving unseen problems after training for 80 steps. Shadow Area: 95\% CI.}
  \label{fig:Q1}
\end{figure}

We evaluate 30 challenging AIME-2025 problems from \citet{balunovic_matharena_2025}. We selected one problem that is unsolvable by Qwen2.5-1.5B Instruct with even 4096 generations. Then, we collect 50 correct solutions from QwQ-32B as the SFT dataset. After SFT for 12 epochs, the accuracy improves to 1.3\% (7/512), which is very small but enough to stabilize RL. Then, we run GRPO on this problem with Rollout \(N\) 512 and learning rate 2e-6. As depicted in \cref{fig:Q1}, after 50 RL steps, the training accuracy improved to 80\%. We also observed a stable response length over 10K tokens, which indicates the model does not simply memorize the final answer.

We further investigate if the model only memorizes solutions without understanding the knowledge behind them. To address this concern, we manually write a solver script and create an additional 8 problems by adjusting parameters in the original problem. If the model only memorizes answers, it is not possible to solve these 8 harder problems. We continue RL on 8 augmented problems without adding SFT data. As a result, the RLed model reached over 80\% accuracy with 80 more steps in \cref{fig:Q1} (Right). Surprisingly, we observed a non-zero accuracy before RL on augmented problems, which indicates the model already learned how to solve augmented problems after training on the original problem.

\subsection{Can We Improve Data Efficiency by Sampling Multiple Solutions?} \label{sec:exp-data-efficiency}

We verify the data scaling effect in a controlled environment. There are two research questions: First, we ask how many solutions are required, at least, to convert a problem from unsolvable to solvable. This experiment is based on Qwen2.5-1.5B Instruct. For 90 problems in the \textit{OMR-seed} dataset, we gradually increase the solution per problem. After each increment, we finetune the model from scratch. We evaluate each SFTed model and record the minimum number of solutions needed to have a non-zero Pass@16. 

\begin{figure}[ht]
  \centering
  \includegraphics[width=0.45\textwidth]{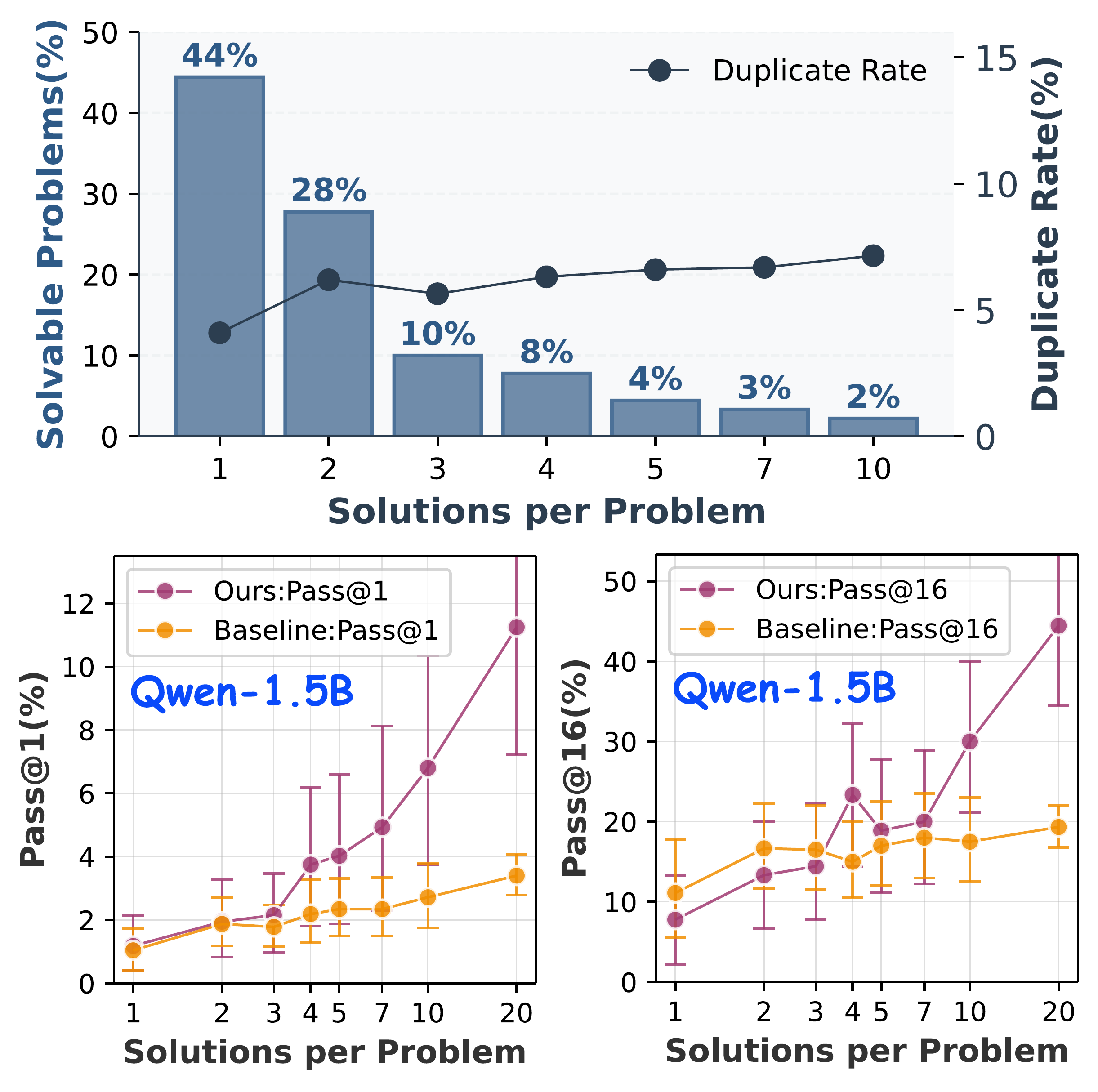}
  \caption{\textbf{Some problems need more solutions to be solvable}. \textbf{Top}: Long-tailed distribution of minimal required solutions per problem. \textbf{Bottom}: Scaling up the solution per problem improves both Pass@1 and Pass@16, which is more effective than scaling up the dataset size. This experiment is based on Qwen2.5-1.5B Instruct. Error bar: 95\% CI.}
  \label{fig:data-eff}
\end{figure}

In \cref{fig:data-eff} (Top), we plot the long-tailed distribution of minimal required solutions. We found that LRM needs more than one solution to learn hard problems. It aligns with the intuition that more difficult problems may contain more complex logic, so that only one solution may not let LRM understand the underlying logic. We also investigate if training on multiple solutions elicits memorization. We computed the duplicate rate, which is defined as the proportion of exactly overlapped tokens between the output trace and any traces in the training set. The duplicate rates for all SFTed models are under 10\%.

Second, we investigate whether scaling the solution per problem improves RL potential as effectively as scaling up the dataset size. We measure the RL potential by observing both Pass@1 and Pass@16 accuracy. For \texttt{ours}, we scale up the solution per problem on the \textit{OMR-seed} dataset. For each SFTed model, we evaluate the Pass@1 and Pass@16 on 90 training problems. For \texttt{baseline}, we sample one solution per problem and enlarge the dataset size up to 1800 problems, and randomly select 200 problems for testing. We combine the \textit{OMR-seed} and \textit{OMR-rest} datasets to get more problems. Since \textit{OMR-seed} is randomly selected from \textit{OMR-rest}, this combination does not change the problem distribution.

In \cref{fig:data-eff} (Bottom), we found that scaling up the solution per problem has a significant effect on seen problems. For example, \texttt{baseline} improves Pass@1 to 3.8\% on 1800 problems, but \texttt{ours} improves Pass@1 to 11.5\% on 90 problems. While scaling up dataset size indeed has a wider generalization effect, the accuracy improvement is also slower, which prevents LRM from learning hard problems when there is no sufficient problem set.

\subsection{Does Our Data Scaling Strategy Work on Larger Language Models?} \label{sec:exp-data-aug}

We investigate whether our data scaling strategy works on larger Language Models. We choose two popular Language Models: Llama3.2-3B Instruct and Qwen2.5-7B Instruct. We follow the same training process as finetuning the model in \cref{fig:data-eff} (Bottom). 

\begin{figure}[ht]
  \centering
  \includegraphics[width=0.45\textwidth]{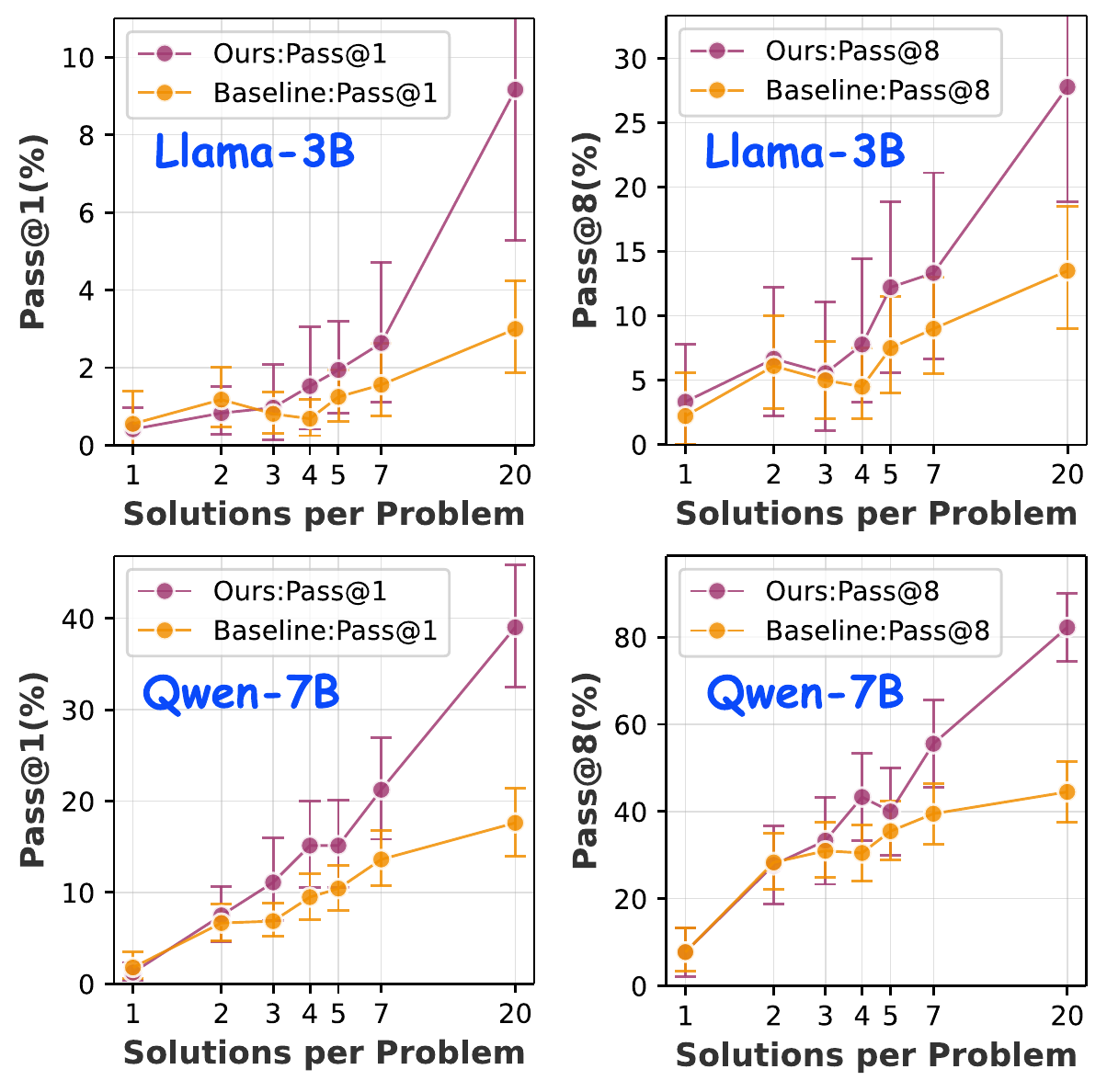}
  \caption{\textbf{Our data scaling strategy also works on Llama3.2-3B Instruct and Qwen2.5-7B Instruct}. Our scaling strategy has significant improvement compared with baseline. Error bar: 95\% CI}
  \label{fig:data-eff-aug}
\end{figure}

In \cref{fig:data-eff-aug}, we found that our data scaling strategy consistently improves on these two Language Models. When a problem has more than 5 solutions, the improvement becomes significant. This indicates that our data scaling strategy is a general method that can be applied to different Language Models. Combined with \cref{fig:data-eff} (Bottom), our data scaling strategy is effective on 1.5B, 3B, and 7B Language Models. Remarkably, the improvement does not shrink when the model size increases from 1.5B to 7B. 

\subsection{Can We Improve Computation Efficiency by Adjusting Rollout \(N\)?} \label{sec:exp-compute-efficiency}

Our method has both positive and negative effects on computational efficiency. The advantage is that we assign different hyperparameters for each subset, so that the overall efficiency may be closer to the optimal point. However, RL in groups prevents learning from other subsets. We investigate whether our method can indeed improve computational efficiency in controlled settings. 

We design a controlled experiment on Qwen2.5-1.5B Instruct. Starting from an SFTed model, we keep total compute and all other hyperparameters identical across experiments and vary only the Rollout \(N\). \texttt{baseline} runs a single RL process on the full set with fixed rollout \(N=80\). \texttt{ours} partitions the 10 problems into 4 difficulty-based subsets and runs independent RL for each subset with fixed rollout \(N\): We use \(N=40\) for easier subsets and \(N=160\) for harder ones. Both experiments use the same learning rate and other settings. We report Pass@16 as a function of the cumulative generated solutions to quantify computation efficiency. We evaluate Pass@16 because Pass@1 is easily influenced by simpler problems, but can not reflect the accuracy on convergence.

\begin{figure}[ht]
  \centering
  \includegraphics[width=0.45\textwidth]{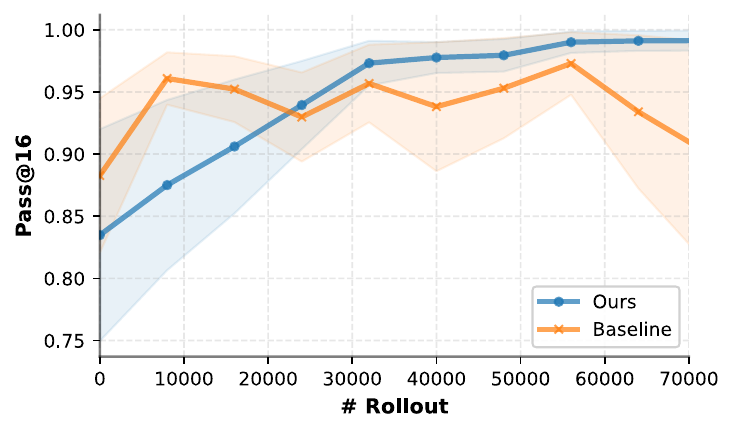}
  \caption{Our method has better computational efficiency. Given equal computation (number of rollouts), our method has a better Pass@16 during RL. Shadow Area: 95\% CI.}
  \label{fig:compute-eff}
\end{figure}

In \cref{fig:compute-eff}, our method demonstrates better Pass@16 and a more stable accuracy curve. We found that the performance gap is mainly from 2 hard problems that were failed in \texttt{baseline}. Overall, by adjusting Rollout \(N\), it is possible to achieve similar or even better performance with the same or even less computation.

\section{Conclusion}

In this paper, we introduce an efficient post-training scaling strategy: \ourmethod. The core idea of our method is to simultaneously scale up data and computation. We found that small LRM can also output stable long reasoning traces by a proper scaling method. Our experiments show that \ourmethod\ is able to train a 0.5B Language Model on problems that are out of the ability boundary. We obtained a significant performance improvement on MATH-500 and the other three math reasoning benchmarks. We hope that our work helps to understand the underlying mechanism of LRM post-training.

\section{Acknowledgement}

This paper is supported by Noncommunicable Chronic Diseases-National Science and Technology Major Project (2024ZD0522702).

\bibliography{HybridReasoning}

\bibliographystyle{icml2026}

\def\subsectioncrefname{Appendix}

\newpage
\appendix
\onecolumn

\section{Proof of Computation Efficiency} \label{sec:appendix-proof-comp-optimal}

\subsection{Proof of SGD Efficiency}

We leverage Stochastic Differential Equations (SDEs) to derive computational efficiency in Reinforcement Learning. We start from the SGD optimizer as a basic case, then generalize to the Adam optimizer \citep{kingma_adam_2017}. To describe the effect of noise, we assume the estimated gradient has a gaussian noise. Let \(\hat g_L = g_L + \epsilon\) to be the estimated gradient, where \(\epsilon \sim \mathcal{N}(0, \frac{1}{N}\Sigma)\). \(N\) is the \textit{Rollout N}, or rollout size per problem. The parameter update rule can be represented as \cref{equ:sgd-update}, where \(g_L=\nabla L(\theta)\) is the gradient.

\begin{equation} \label{equ:sgd-update}
  \Delta \theta_k = - \eta \hat g_L = - \eta g_L - \eta \epsilon
\end{equation}

When the learning rate is low enough, we approximate the discrete gradient descent as a continuous process in \cref{equ:sgd-sde}, where \(W_t\) is the standard Brownian motion.

\begin{equation} \label{equ:sgd-sde}
  d\theta_t = - \eta g_L \, dt + \frac{\eta}{\sqrt{N}} \Sigma^{1/2} \, dW_t
\end{equation}

By It\^o Lemma, we can describe how fast the accuracy increases by computing the drift term of reward \(R(\theta)=\mathbb{E}_{a\sim \pi_\theta, s}[r(a,s)]\), where \(r(a,s)\) is a 0-1 binary reward for each solution sequence. Let \(g_R = -g_L = \nabla R(\theta)\) and  \(H_R(\theta) = \nabla^2R(\theta)\) to be the Hessian matrix. We have:

\begin{align}
  dR(\theta_t) & = \nabla R(\theta_t)^\top d\theta_t + \frac{1}{2} \text{tr}(H_R(\theta_t) d\theta_t d\theta_t^\top) \notag \\
  & = \left[ \eta g_R^\top g_R + \frac{1}{2}\frac{\eta^2}{N}\text{tr}(H_R \Sigma)\right] dt + g_R^\top \frac{\eta}{\sqrt{N}}\Sigma^{1/2} dW_t \notag
\end{align}

We derive the computational efficiency by dividing the increasing speed of accuracy by Rollout \(N\). In \cref{equ:sgd-efficiency}, we represent it in the form of a quadratic equation.

\begin{align} \label{equ:sgd-efficiency}
  \mathcal{E}_{\text{SGD}} & = \frac{1}{N} \frac{d}{dt}\mathbb{E}[R(\theta_t)] \notag \\
  & = \frac{\eta}{N} g_R^\top g_R + \frac{1}{2} \left(\frac{\eta}{N}\right)^2 \text{tr}(H_R \Sigma)
\end{align}

We found that \(\mathcal{E}_{\text{SGD}}\) is related to \(\eta/N\) as a whole, which means adjusting \(\eta\) or \(N\) have the same effect on computational efficiency: They both control \(\mathcal{E}_{\text{SGD}}\) by adjusting \(\eta/N\). We can write \cref{equ:sgd-efficiency} in a simplified form as \cref{equ:sgd-efficiency-final}.

\begin{align}
  \mathcal{E}_{\text{SGD}} &= \frac{\eta}{N}\mu_p -  \left ( \frac{\eta}{N} \right )^2 \mu_n \label{equ:sgd-efficiency-final}\\
  \text{where } \mu_p & = g_R^\top g_R, \quad \mu_n = -\frac{1}{2}\text{tr}(H_R \Sigma) \notag
\end{align}

\subsection{Proof of Adam Efficiency}

For the Adam optimizer, we introduce some assumptions to derive a simplified approximation. We assume the exponential moving average in Adam can be approximated as a slice window average. The window length is defined as \(N_m := 1/(1-\beta_1)\) for first moment and \(N_v := 1/(1-\beta_2)\) for second moment. We use \(g_L(k) \) to denote the gradient in step \(k\). We further assume the gradient change is slow enough and the noises in each step are independent, so that \( \mathbb{E}[g_L(k)] \approx g_L\). Therefore, the first moment \(m(k)\) and the second moment \(v(k)\) can be approximated as \cref{equ:adam-approx-m} and \cref{equ:adam-approx-v}. 

\begin{align} 
  m(k) & \approx \frac{1}{N_m} \sum_{k=1}^{N_m}g_L(k) \sim \mathcal{N}(g_L, \frac{1}{N_m}\frac{\Sigma}{N}) \label{equ:adam-approx-m}\\
  v(k) & \approx \frac{1}{N_v} \sum_{k=1}^{N_v}[g_L(k)^2] \sim \mathcal{N}\left( g_L^2 + \text{diag}\left(\frac{\Sigma}{N}\right), \frac{1}{N_v} \left( 2 \left(\frac{\Sigma}{N}\right)^2 + 4 (g_L g_L^T) \left(\frac{\Sigma}{N}\right) \right) \right) \label{equ:adam-approx-v}
\end{align}

Note that \(\Sigma^2\) in \cref{equ:adam-approx-v} is an in-element operation. In general case, we have \(\beta_1=0.9\) and \(\beta_2=0.999\). Therefore, \(N_m=10\) and \(N_v=1000\). Since Adam is often used with learning rate warmup, the first moment has a small dependency on the second moment. We analyze this dependency by decomposing \(v(k)\) into two parts. We can write \(v(k) = v_1(k) + v_2(k)\) and \(v_2(k)\) is independent of both \(m\) and \(v_1(k)\):

\begin{align}
  v_1(k) = \frac{1}{N_v} \sum_{k=1}^{N_m}[g_L(k)^2] \notag \\
  v_2(k) = \frac{1}{N_v} \sum_{k=N_m}^{N_v}[g_L(k)^2] \label{equ:adam-approx-v12}
\end{align}

Here we introduce a lemma from \citet{li_stochastic_2019} (Lemma 1) and \citet{compagnoni_adaptive_2025} (Lemma C.6) to derive a 1-order SDE approximation.

\begin{lemma} \label{lemma:adam-sde}
Let \(0 < \eta < 1\). Consider a stochastic process \(X_t, t\geq 0\) satisfying an SDE:
\begin{equation}
  dX_t = b(X_t) dt + \sqrt{\eta}\sigma(X_t) dW_t
\end{equation}
with \(X_0=x \sim \mathbb{R}^d\) and \(b,\sigma\) together with their derivatives belong to \(G\). Define the one-step different \(\Delta = X_\eta - x\), and indicate the i-th component of \(\Delta\) as \(\Delta_i\). We have:

\begin{align}
  \mathbb{E}[\Delta_i] & = b_i(x) \eta + \frac{1}{2}[\sum^d_{j=1}b_j\partial_{e_j}b_i] + O(\eta^2) \notag \\
  \mathbb{E}[\Delta_i \Delta_j] & = [b_ib_j + \sigma\sigma^T_{(ij)}]\eta^2 + O(\eta^3) \notag \\
  \mathbb{E}[\prod^s_{j=1}\Delta_{i_j}] & = O(\eta^3) \text{ for all } s \geq 3, i_j= 1,...,d.
\end{align}
\end{lemma}

According to \cref{lemma:adam-sde}, we can derive a 1-order SDE by computing the first and second moments of the parameter update rule \cref{equ:adam-update}.

\begin{equation} \label{equ:adam-update}
  \Delta \theta_k = - \eta \frac{m}{\sqrt{v}+\epsilon}
\end{equation}

The first moment is derived in \cref{equ:adam-first-moment}. We expand \(\mathbb{E}[1/\sqrt{v}]\) to the second order at \(v=g_L^2\). For simplicity, we denote the i-th element in \(g_L\) as \(g\) and the i-th element in \(\text{diag}(\Sigma^{1/2}/\sqrt{N})\) as \(\sigma\). First, we compute some necessary expectations:

\begin{align}
  \mathbb{E}[g_L(k)^3] & = g^3 + 3g \sigma^2 \\
  \mathbb{E}[mv_1] & = \mathbb{E}[\frac{1}{N_m}\sum_{k=1}^{N_m}g_L(k) \cdot \frac{1}{N_v}\sum_{j=1}^{N_m}g_L(j)^2] \notag \\
  & = \frac{1}{N_m N_v} \mathbb{E} \left[ (N_m^2 - N_m)g(g^2 + \sigma^2) + N_m (g^3+3g \sigma^2) \right] \notag \\
  & = \frac{N_m}{N_v}g(g^2 + \sigma^2) + 2(\frac{N_m}{N_v})^2g \sigma^2\\
  \mathbb{E}[m v_2] & = (1 - \frac{N_m}{N_v})g(g^2 + \sigma^2)
\end{align}

We can use the above results to compute \(\mathbb{E}[m(v_1+v_2-g^2)]\) and \(\mathbb{E}[m(v_1+v_2-g^2)^2]\):

\begin{align}
  \mathbb{E}[m(v_1+v_2-g^2)] & = \mathbb{E}[m v_1] + \mathbb{E}[m v_2] - g^3 \notag \\
  &= \left[1 + 2\left(\frac{N_m}{N_v}\right)^2\right] g\sigma^2  \\
  \mathbb{E}[m(v_1+v_2-g^2)^2] & \approx (1+\frac{N_m}{N_v})^2 g \sigma^4 
\end{align}

Then, we approximate \(\mathbb{E}[\Delta \theta_i]\) by Taylor expansion and plug in the above expectations:

\begin{align}
  \mathbb{E}[\Delta \theta_i] & \approx -\eta \mathbb{E}[m/\sqrt{v}] \notag\\
  & \approx -\eta \mathbb{E}\left[ m|g|^{-1}-\frac{1}{2}m|g|^{-3}(v_1+v_2-g^2)+\frac{3}{8}m|g|^{-5}(v_1+v_2-g^2)^2 + \mathcal{O}(m(v_1+v_2-g^2)^3/|g|^7)\right] \label{equ:adam-first-moment} \notag\\
  & \approx -\eta\; \text{sign}(g) \left[ 1 - \frac{1}{2}(1+2\frac{N_m}{N_v})(\frac{\sigma}{g})^2 + \frac{3}{8}(1+\frac{N_m}{N_v})^2(\frac{\sigma}{g})^4  + \mathcal{O}((\sigma/g)^6) \right] 
\end{align}

The first moment expectation can be simplified as \cref{equ:adam-first-moment-simplified}.

\begin{equation} \label{equ:adam-first-moment-simplified}
  \mathbb{E}[\Delta \theta_i] \approx -\eta\; \text{sign}(g) \left[ 1 - \frac{1}{2}(\frac{\sigma}{g})^2 + \frac{3}{8}(\frac{\sigma}{g})^4 + \mathcal{O}((\sigma/g)^6) \right]
\end{equation}

The second moment is derived in \cref{equ:adam-second-moment} with a similar approximation. 

\begin{align}
  \mathbb{E}[(\Delta \theta_i)^2] & \approx \eta^2 \mathbb{E}[m^2/v] \notag \\
  & \approx \eta^2 (\frac{1}{N_m}\sigma^2 + g^2) \left[\frac{1}{g^2} - \frac{1}{g^4}\sigma^2+\frac{1}{g^6}(\sigma^4 + \frac{4}{N_v}\sigma^2g^2) + \mathcal{O}(\sigma^6/g^8)  \right] \notag \\
  & \approx \eta^2\left[1 + (1/N_m-1+4/N_v)(\frac{\sigma}{g})^2 + (1-1/N_m(1-4/N_v))(\frac{\sigma}{g})^4+1/N_m(\frac{\sigma}{g})^6 \right] + \eta^2 \mathcal{O}((\sigma/g)^6 ) \notag \\
  & \approx \eta^2\left[1 + (1/N_m-1)(\frac{\sigma}{g})^2 + (1-1/N_m)(\frac{\sigma}{g})^4+1/N_m(\frac{\sigma}{g})^6 \right] + \eta^2 \mathcal{O}((\sigma/g)^6 ) \label{equ:adam-second-moment}
\end{align}

Let's denote \(x=(\sigma/g)^2\), we have \(x \propto 1/N\). From \cref{lemma:adam-sde}, the 1-order SDE approximation is derived in \cref{equ:adam-sde}.

\begin{align}
  b[X_t]&=\mathbb{E}[\Delta \theta_i]/\eta=-\text{sign}(g)(1-\frac{1}{2}x+\frac{3}{8}x^2 + \mathcal{O}(x^3)) \notag \\
  \sigma^2(X_t)&=\mathbb{E}[(\Delta \theta_i)^2]/\eta^2-b[X_t]^\top b[X_t] \notag\\
  & =\frac{1}{N_m}x - \frac{1}{N_m}x^2 + \left( \frac{3}{8} + \frac{1}{N_m}\right) x^3 + \mathcal{O}(x^4) \notag\\
  \notag \\
  d\theta_t & = b[X_t] dt + \sqrt{\eta}\sigma(X_t) dW_t \notag \\
  & = \eta b[X_t] dt + \eta \sigma(X_t) dW_t \label{equ:adam-sde}
\end{align}

Now we can compute the computational efficiency of the Adam optimizer in \cref{equ:adam-efficiency}. We only keep the highest order of \(x\) in each term. 

\begin{align} 
  \mathcal{E}_{\text{Adam}} & = \frac{1}{N} \frac{d}{dt}\mathbb{E}[R(\theta_t)] \notag \\
  & = \frac{\eta}{N}g_R^\top b[X_t] + \frac{1}{2}\frac{\eta^2}{N}\text{tr}(H_R \sigma^2(X_t)) \notag \\
  & \approx \frac{\eta}{N}g_R^\top \text{sign}(g_R) \left[ 1 - \frac{1}{2}x + \frac{3}{8}x^2 + \mathcal{O}(x^3)\right] + \frac{1}{2}\frac{\eta^2}{N}\text{tr}\left(H_R \left[ \frac{1}{N_m}x - \frac{1}{N_m}x^2 + \mathcal{O}(x^3) \right]\right) \notag \\
  & \approx \frac{\eta}{N}g_R^\top \text{sign}(g_R) + \frac{1}{2 N_m g_R^\top g_R}\left(\frac{\eta}{N}\right)^2\text{tr}(H_R \Sigma) \label{equ:adam-efficiency}
\end{align}

We can ignore the high order of \(x\) when \(x=(\sigma/g)^2\) is small enough, which requires \(N\) to be large enough. From \cref{equ:adam-efficiency}, we found that Adam's computational efficiency has the same form as SGD's efficiency in \cref{equ:sgd-efficiency}. Therefore, adjusting \(\eta\) or \(N\) has the same effect on computational efficiency for the Adam optimizer. 

\section{Implementation Details}

\subsection{Algorithm Description} \label{appendix:alg-desc}

We describe the pseudocode of \ourmethod\ in \cref{alg:ours}.

\begin{algorithm}[t]
\begin{small}
\caption{The implementation of \ourmethod}
\label{alg:ours}
\begin{algorithmic}[1]
\STATE \textbf{Initialize} Policy $\pi_0$, Problem Pools $\mathcal{U}$(Unsolvable), $\mathcal{S}$(Solvable),$\mathcal{D}$(Solved),$\mathcal{F}$(Failed). All problems start in $\mathcal{U}$.
\STATE \textbf{Initialize} Thresholds: $\tau_{\text{SFT}}\gets5\%$, $\tau_{\text{RL}}\gets70\%$
\STATE Perform warmup SFT for deep-thinking mode output

\FOR{$k = 0,1,2,\dots$}
    \STATE \textbf{----------- SFT Stage -----------}
    \STATE Add $2^k-1$ new solutions for each $p \in \mathcal{U}$, then finetune $\pi$ on new solutions
    \STATE Move all $p$ from $\mathcal{U}$ to $\mathcal{S}$ if Accuracy$(p) \ge \tau_{\text{SFT}}$
    
    \STATE \textbf{----------- RL Stage -----------}
    \STATE Partition $\mathcal{S}$ into accuracy-sorted groups $\{\mathcal{G}_1,\dots,\mathcal{G}_m\}$
    \FORALL{$\mathcal{G}_i$}
        \STATE Train $\pi$ with budget $N_i$
        \IF{$\mathcal{G}_i$ fails}
            \STATE Move all problems in $\mathcal{G}_i$ to $\mathcal{F}$ if budget exhausted, else double $N_i$ and retry
        \ELSE
            \FORALL{$p \in \mathcal{G}_i$}
                \IF{Accuracy$(p) \ge \tau_{\text{RL}}$}
                    \STATE Move $p$ to $\mathcal{D}$
                \ELSIF{Accuracy$(p) \le \tau_{\text{SFT}}$}
                    \STATE Move $p$ to $\mathcal{U}$
                \ELSE
                    \STATE Move $p$ to $\mathcal{S}$
                \ENDIF
            \ENDFOR
        \ENDIF
    \ENDFOR
    
    \STATE \textbf{----------- Re-Distillation -----------}
    \STATE For each $p \in \mathcal{S} \cup \mathcal{D}$, collect latest $\sim 100$ correct RL solutions into $\mathcal{B}_p$
    \STATE Finetune $\pi$ on $\bigcup_p \mathcal{B}_p$
\ENDFOR
\STATE \textbf{----------- Training Target Policy -----------}
\STATE Collect all latest $\sim 100$ correct RL solutions per problem into $\mathcal{B}_p$
\STATE Collect all SFT solutions into $\mathcal{S}_p$
\STATE Finetune $\pi_0$ on $\bigcup_p \mathcal{S}_p$ first, then $\bigcup_p \mathcal{B}_p$ for final policy
\end{algorithmic}
\end{small}
\end{algorithm}

\subsection{Practical Algorithm for Maximizing Computation Efficiency (Discussed in \cref{sec:takeaways})} \label{sec:practical-alg}

As discussed in \cref{sec:takeaways}, it is nearly impossible to adjust \(\eta/N\) for each problem practically. When the RL dataset contains multiple problems, their gradients are mixed in the policy update. \ourmethod\ made some approximations to improve computational efficiency in a practical method.

Our approach is keeping \(\eta\) fixed and performing a linear search for \(N\). Specifically, we sort problems according to estimated accuracy and only keep solvable problems in the RL training set. Then, we divide all problems into groups. The group with lower accuracy has a smaller Rollout \(N\) at the start. For each group, we enlarge Rollout \(N\) by 2 times if it failed. We manually double \(N\) when the accuracy stays on a plateau, which is a simple but effective strategy for real-world tasks.

\subsection{Algorithm Adaptation: Importance Sampling between vLLM and FSDP (Discussed in \cref{sec:exp-settings})} \label{sec:rl-importance-sampling}

Our training algorithm is based on GRPO. We further add a token-level importance sampling between vLLM \citep{kwon_efficient_2023} and the FSDP engine, which mitigates the distribution shift due to implementation differences. We found that the instability mainly comes from punishing incorrect tokens, which are overly sampled by vLLM. For example, vLLM may generate a few tokens that should not appear with such high frequencies, but are being used in training. We solve this problem by assigning a near-zero advantage to those tokens.

Specifically, the re-weighted advantage is defined in \cref{equ:adv}. The vanilla advantage function \(\hat{A}_{i,t}\) is computed by GRPO.

\begin{equation} \label{equ:adv}
   \hat{A}^{IS}_{i,t} = \min \left[ \frac{\pi_{\theta}^\text{FSDP}(o_{i,t} \mid q, o_{i,< t})}{\pi_{\theta}^\text{vLLM}(o_{i,t} \mid q, o_{i,< t})}, 1+\epsilon^{IS} \right] \hat{A}_{i,t}
\end{equation}

In the final loss function \cref{equ:loss-func}, we removed GRPO's ratio clip because we only use one mini-batch for each policy update. Hence, the ratio between the old policy and the current policy is always 1.0. We also removed the KL divergence loss and entropy loss for simplicity.

\begin{equation} \label{equ:loss-func}
   \nabla_\theta \mathcal{L}_{\text{GRPO}}(\theta) = - \frac{1}{G} \sum_{i=1}^G \frac{1}{|o_i|} \sum_{t=1}^{|o_i|} \left[ \hat{A}^{IS}_{i,t} \nabla_\theta \log \pi_{\theta}^\text{FSDP}(o_{i,t} \mid q, o_{i,< t}) \right]
\end{equation}

\section{Experiment Details} \label{sec:appendix-exp-details}

We listed important experimental settings in \cref{sec:exp-settings}. Most RL experiments are based on VeRL \citep{sheng_hybridflow_2024}. In this section, we provide hyperparameters, training details, and additional experimental settings.

\subsection{Experimental Details of \ourmethod\ (Discussed in \cref{sec:exp-large-scale})}

\textbf{Warmup SFT and Model Initialization}: We first train Qwen2.5-0.5B Instruct to output stable long reasoning traces. To achieve this, we finetune Qwen2.5-0.5B Instruct on 920 solutions of 10 selected problems from \textit{OMR-seed}. The solutions are generated by QwQ-32B. We use a learning rate of 1e-5, batch size 4, and 16 epochs in warmup SFT. The SFTed model is able to output long reasoning traces, but often fails to generate an end-of-sentence token within the given max token limit. We further perform RL on the same 10 problems for 60 steps. We use the RLed model as the initial policy in our main pipeline. We keep the warmup dataset very small to only adjust the output mode, but not inject too much knowledge. Additionally, we found that this policy does not outperform Qwen2.5-0.5B Instruct, despite it thinking with more tokens.

\textbf{Iterative Training}: The overall settings are described in \cref{tab:exp-settings}. The training process contains 3 iterations. In iteration 0, there is no SFT stage because the initial policy can solve some problems without additional SFT. To measure solvable problems, we generate 20 solutions for each problem in the RL training set. For those  problems with accuracy over 5\%, we put them into Solvable set and sort by accuracy, then divide them into multiple subsets. Each subset contains 100 problems. Note that each subset runs RL independently.

\begin{table}[ht]
\begin{small}
\centering
\caption{Experimental Settings for Iterative Training}
\label{tab:exp-settings}
\setlength{\tabcolsep}{4pt}
\begin{tabularx}{\textwidth}{p{2.2cm} X X}
\toprule
\textbf{Stage} & \textbf{Training Settings} & \textbf{Data / Other Settings} \\
\midrule

\textbf{Warmup SFT} &
\makecell[l]{
-- Learning rate: $1\times10^{-5}$ \\
-- Batch size: 4 \\
-- Epochs: 16
} &
\makecell[l]{
-- Dataset: 920 correct solutions by QwQ-32B \\
-- 10 problems from OpenMathReasoning \\
-- Base model: Qwen2.5-0.5B Instruct
} \\

\midrule
\makecell[l]{
\textbf{Warmup RL} \\ \textbf{(60 steps)}
} &
\makecell[l]{
-- Learning rate: $2\times10^{-6}$ \\
-- Train batch size: 20 \\
-- Rollout \(N\): 80 \\
-- Policy update batch: Train batch $\times$ Rollout \(N\)
} &
\makecell[l]{
-- Temperature: 0.7; Top-p: 1.0 \\
-- Max response tokens: 16000 \\
-- Adam: $\beta_1=0.9, \beta_2=0.999$ \\
-- Dataset: 10 problems \\
-- Removed: ratio clip, KL loss, reference model
} \\

\midrule
\textbf{Evaluation Stage} &
\makecell[l]{
-- Generate 20 solutions per problem
} &
\makecell[l]{
-- Same decoding settings as Warmup RL \\
-- Accuracy $>5\%$ labeled solvable
} \\

\midrule
\textbf{SFT Stage} &
\makecell[l]{
-- Learning rate: $1\times10^{-5}$ \\
-- Batch size: 16
} &
\makecell[l]{
-- Epochs: 12 \\
-- Base model: last iteration
} \\

\midrule
\textbf{RL Stage} &
\makecell[l]{
-- Train batch size: 20--40 \\
-- Rollout \(N\): 20--40
} &
\makecell[l]{
-- Manually adjust \(N\) on plateau \\
-- Other settings same as Warmup RL
} \\

\midrule
\textbf{Re-distill Stage} &
\makecell[l]{
-- Learning rate: $1\times10^{-5}$ \\
-- Batch size: 16
} &
\makecell[l]{
-- Epochs: 4 \\
-- 100 latest correct solutions per problem
} \\

\midrule
\makecell[l]{
\textbf{Target Policy} \\ \textbf{(SFT)}
 } &
\makecell[l]{
-- Base model: initial policy
} &
\makecell[l]{
-- SFT on 30K correct solutions (12 epochs) \\
-- Re-distill on 98K replay buffer solutions (4 epochs)
} \\

\bottomrule
\end{tabularx}
\end{small}
\end{table}

In the RL stage, we use a learning rate of 2e-6, temperature 0.7, and top-p 1.0. The train batch size and Rollout \(N\) are varied across different subsets. As an initial setting, we use a train batch size of 40 and Rollout \(N\) 20 for easier problems, and batch size 20 with Rollout \(N\) 40 for harder problems. We manually adjust the train batch size and Rollout \(N\) when RL reaches a plateau or fails. In the policy update phase, the model weights always update once. Therefore, we removed the ratio clip because it has no effect. We also removed the KL constraint and the reference model for simplicity. 

In iteration 0, 273 out of 4710 problems become solvable. We divided 273 problems into 5 subsets according to the accuracy. After RL, 178 problems are solved (i.e., accuracy over 70\%), 57 problems still have less than 5\% accuracy. We re-distill 216 problems into the next iteration. In Re-distillation stage, we collect 100 correct solutions from the latest replay buffer and SFT for 4 epochs. The re-distillation is performed with a learning rate of 1e-5 and a batch size of 16.

In iteration 1, we collect 2 correct responses to finetune the redistilled policy from iteration 0. After finetuning on 8764 solutions, there are 195 problems becomes solvable. After RL on 402 solvable problems, 333 of them entered re-distillation. In iteration 2, we collect 8 solutions for the unsolvable problem, which lead to 22.6K solutions generated from QwQ-32B. After SFT, there are 1291 problems out of 4710 problems that become solvable.

\textbf{Training Target Policy}: After iteration 2, we train the target policy from the warmup policy to avoid catastrophic forgetting. We first finetune the initial policy on all SFT samples (30K from iteration 1 and 2). Then continue finetuning on 98K re-distillation samples.

\subsection{Experimental Details of Baselines (Discussed in \cref{sec:baseline-settings})} \label{sec:appendix-baseline-details}

\textbf{SFT w/o data scaling}: All baseline models are based on this SFTed model. We finetune Qwen2.5-0.5B Instruct with 32K problem-solution pairs sampled from the OpenMathReasoning dataset \citep{moshkov_aimo-2_2025}. We first collect all 4710 problems in \textit{OMR-rest} dataset (see \cref{sec:exp-settings}) and randomly choose one correct solution. Then, we extend the SFT dataset by randomly collecting more combinatorial and probabilistic problems in OpenMathReasoning. This aligns with most existing methods, which create an extensive dataset but with only one solution per problem.

\textbf{COMPASS}: We follow the settings in \citet{chalumeau_combinatorial_2024}. The latent space for sampling the compass vector is a 16-dimensional box with [-1, 1] bounds. The compass vector is multiplied by a learnable linear layer, then added to the hidden state of the first valid token. The linear layer is set to all zeros at the initial state. In the training phase, we generate 8 candidates and select the best one. The decoding setting is temperature 0.7, top-p 1.0, and max new token 16000. The policy is updated by a learning rate of 2e-6 and one mini-batch for each step. In the inference phase, we use Covariance Matrix Adaptation Evolution Strategy (CMA-ES) to sample candidates. We sample 10 candidates for each problem and repeat for at most 50 times to find optimal compass vectors. While this method requires checking the answer during inference, we only sample final solutions from the searched compass vectors, which reduces the effect of random guessing. Since the implementation of COMPASS is about 20x slower than VeRL, we use the RLed model from SAPO and train COMPASS for fewer steps. 

\textbf{Speed-Control}: We first prompt the RLed model from Scale-RL with a `To' prefix to obtain shrot response, then use normal decoding settings to obtain a long response. The construction of the speed control vector is done on the MATH training set with 11K problems. After filtering, we use about 3k correct pairs to perform PCA. Each short response is truncated to preserve the first 2 steps (each step ends with $\backslash$n$\backslash$n). The long response is truncated with the nearest token length. In generation, we use a constant intervention and inject the PCA vector for all layers. We leverage the code in \citet{lin_controlling_2025} to apply the PCA vector in inference.

\textbf{SAPO \& GRPO}: We implement SAPO \citep{gao_soft_2025} with \(\tau_{\text{pos}}=1.0\) and \(\tau_{\text{neg}}=1.05\). In all RL experiments, we only optimize once for each policy update. So \(r_{i,t}(\theta)=\pi_\theta/\pi_{\theta_{\text{old}}}=1\) always hold. In SAPO, GRPO and Scale-RL, we use temperature 0.7, top-p 1.0, Rollout \(N\) 32, and train batch size 128.

\textbf{Scale-RL}: We follow the best practice in \citet{khatri_art_2025} to implement Scale-RL. Specifically, we implemented CISPO (without length penalty), advantage batch norm, FP32 for LM head, zero std rollout filter, and an acc over 90\% samples filter. For CISPO, we removed the length penalty to maximize performance improvement. For the advantage batch norm, we first remove all zero std rewards, then compute the mean and std based on the rest samples. We did not implement one-step off-policy RL because it does not affect the performance. 

\subsection{Performance Evaluation Details} \label{sec:appendix-eval-details}

Here we list all benchmark settings and evaluation details. For all benchmarks, we do not use tool calling or few-shot examples. We also applied a simple decoding setting: temperature 0.7 and top-p 0.95. The max token length is set to 25K to avoid truncation.

\textbf{MATH-500}: We selected 54 combinatorial and probabilistic problems out of 500 problems. The selection is automatically performed by prompting Qwen2.5-32B. We duplicate the test set 16 times to achieve stable performance.

\textbf{OlymMATH}: We collected all 29 problems, which are labeled as Combinatorics, from 100 problems in OlymMATH-EN-EASY. We duplicate the test set 128 times.

\textbf{AMC}: We selected 19 problems from a combined problem set of AMC12-2022 and AMC12-2023. The problem selection is the same as MATH-500. We duplicate the test set 128 times.

\textbf{OpenMathReasoning}: We randomly selected 200 problems as a test set from all filtered problems and duplicated them 8 times.

\textbf{Reasoning GYM}: This dataset is synthesized by a program. We randomly selected 100 tasks and generated 5 problems for each task. We use the default task parameters in problem generation. We found that the default task setting is hard enough for a 0.5B level Language Model.

\section{Ablation Study Details} \label{sec:appendix-ablation}

\subsection{Experimental Details of Single Problem Data Scaling (\cref{sec:Q1})}

There are two sub-experiments in \cref{sec:Q1}. In the first experiment, we aim to demonstrate that post-training can be done even at a very small scale. We first finetune a 1.5B Language Model to achieve about 2\% accuracy on a challenging reasoning problem, then use Reinforcement Learning to achieve over 80\% accuracy with long reasoning traces (over 10K tokens per response).

\textbf{Step1: Problem Selection and Difficulty Verification}: We select all 11 combinatorial and probabilistic problems in AIME-2025 dataset. Then we generate 8,192 solutions by prompting Qwen2.5-1.5B Instruct. The decoding setting is temperature 0.7, top-p 0.95, max response token 16000. We found that no solution has a correct answer for the following problem:

\begin{center}
  \textit{Let $ S $ be the set of vertices of a regular 24-gon. Find the number of ways to draw 12 segments of equal lengths so that each vertex in $ S $ is an endpoint of exactly one of the 12 segments.}
\end{center}

We choose this problem because it exceeded the ability boundary of the foundation model, and it is also suitable to derive related problems, such as replacing \textit{24-gon} with \textit{26-gon} or others. 

\textbf{Step2: SFT with 50 solutions}: To collect 50 correct solutions, we prompt QwQ-32B and filter incorrect solutions to create the SFT dataset. Then we finetune Qwen2.5-1.5B Instruct with learning rate 1e-5, batch size 8, and 16 epochs. We use Adam optimizer with \(\beta_1=0.9, \beta_2=0.95\). 

\textbf{Step3: RL on one single problem}: After SFT, the accuracy improves to about 2\%. We perform RL on just one problem with Rollout \(N\) 512, learning rate 2e-6, and temperature 0.7. The top-p is set to 1.0. We use Adam optimizer with \(\beta_1=0.9, \beta_2=0.999\). 

In the second experiment, we aim to verify if RL trained policy really learned the knowledge rather than just memorized solutions. 

\textbf{Step4: Generate augmented problems}: We write a python solver to generate solutions for \textit{M-gon} cases. We set \(M=26,28,...,40\) to generate an additional 8 problems. Since solving a larger M is non-trivial, the model can only solve these problems by learning a generalized methodology. We did not generate any solutions from QwQ-32B for new problems.

\textbf{Step5: RL on augmented problems}: We use the same RL setting except Rollout \(N\) and batch size. In augmented RL, the train batch size is 8, and Rollout \(N\) is 64. Each policy update requires 512 samples.

\subsection{Experimental Details of Data Efficiency (\cref{sec:exp-data-efficiency})}

\subsubsection{Minimal Required Solutions to Make a Problem Solvable (\cref{fig:data-eff} Top)} \label{sec:exp-minimal-solutions}

In this experiment, we aim to test how many solutions are required to make a problem solvable. We define \textit{Solvable} here as generating at least one correct answer in 16 randomly generated solutions. 

\textbf{Dataset}: We use 90 problems in \textit{OMR-seed} dataset, which are combinatorial and probabilistic problems. For each problem, we prompt QwQ-32B to generate 10 correct solutions. 

\textbf{SFT Settings}: To scale up the solution per problem, we perform 7 independent SFTs with \(M=1,2,3,4,5,7,10\). For each SFT, we randomly collect \(M\) correct solutions for each problem. For example, we use \(5 \times 90 = 450\) samples from \(M=5\). The SFT hyperparameters remain the same for each SFT. We use an extreme setting to maximize learning effect: learning rate 1e-4, batch size 16, and 12 epochs. Note that each solution needs to generate about 10K tokens before output final result, the model may fail to output the correct answer by purely memorizing solutions. We found that the generalization effect improves as the model overfits the training set, but some problems still need more solutions to become solvable. 

\textbf{Compute Duplicate Rate}: An important concern is that the model may memorize the easiest solution to `learn' a problem. To address this concern, we compute the duplicate rate for each generated solution. The definition of duplicate rate is the highest proportion of prefix tokens that overlapped with any solution in the training set. We use a strict match to detect overlap. For example, if \(M=2\) and a generated solution matched 10\% prefix tokens for one training solution, and 15\% for another training solution, the duplicate rate for this generated solution is 15\%. Then we compute the average rate for all generated solutions in \cref{fig:data-eff} (Top).

\subsubsection{Improve RL Potential with Better Data Scaling (\cref{fig:data-eff} Bottom)}

In this experiment, we investigate if scaling up the solution per problem (\texttt{ours}) is a better strategy than scaling up the dataset size (\texttt{baseline}). We test these two scaling strategies under controlled experiments.

\textbf{Dataset}: For \texttt{ours}, we use the \textit{OMR-seed} dataset, which contains 90 problems with 20 correct solutions per problem. For baseline, we randomly select 1800 problems from \textit{OMR-rest} dataset, so that the total amount of data aligns with our method (90 problems * 20 solutions = 1800 problems * 1 solution). Similarly, we perform 8 independent SFT with \(M=1,2,3,4,5,7,10,20\) for our method, to compare with the baseline's 8 independent SFT on \(90,180,...,1800\) problems.

\textbf{SFT Settings}: The hyperparameters remain the same for each SFT. We use a learning rate of 1e-5, batch size 16, and 12 epochs. This setting aligns with a realistic training scenario, where the model does not significantly lose ability in overfitting new solutions.

\textbf{Eval Settings}: For our method, we test the Pass@K on 90 trained problems. For baseline, we test on trained problems, which vary by \(M\). If the training set is over 200 problems, we randomly choose 200 problems to compute Pass@K. Since this experiment aims to test RL potential, the decoding setting aligns with the RL rollout phase. For each problem, we generate 16 solutions with temperature 1.0, top-p 1.0, and max token length 16K. 

\subsection{Experimental Details of Computational Efficiency (\cref{sec:exp-compute-efficiency})}

In this experiment, we investigate whether adjusting Rollout \(N\) and learning rate helps stabilize RL without more computations. The theoretical analysis in \cref{sec:rl-scaling} shows that it is possible to improve computational efficiency by sticking to the optimal \(\eta/N\) point. However, the practical algorithm is affected by several factors. For example, spliting dataset into subsets improves overall efficiency because each subset has a hyperparameter closer to the optimal point, but it also blocks learning from other subsets. We aim to verify if it is possible to improve efficiency in a real case.

\textbf{Dataset}: We first select 10 unsolvable problems from \textit{OMR-seed} dataset. Each problem has no correct answer after 4,000 rollouts from Qwen2.5-1.5B Instruct. We collected 920 solutions in total for 10 problems to boost Pass@1 accuracy, so that each problem becomes solvable. Each problem has different optimal hyperparameters because they are finetuned with different number of solutions. This simulates the real case in RL training. We further divide the 10 problems into 4 groups according to their initial accuracies.

\textbf{Experimental Settings}: For SFT on 920 solutions, we use learning rate 1e-5, batch size 16 and 12 epochs. For RL, we use a learning rate of 7e-6, temperature 0.7, top-p 1.0 and Adam optimizer with 5 warmup steps. In \texttt{ours}, the Rollout \(N\) is adjusted in different subsets. We use N=40/80/80/160 for four subsets. For simplicity, we did not adjust hyperparameters during training. In \texttt{baseline}, we use a fixed Rollout \(N\) of 80 for all problems. 

For computing efficiency, we sum up all training steps' computations for \texttt{ours} to get total computation. Since each subset converges at different steps, this method is a rough estimate. We found that the two hardest problems in \texttt{ours} have a significantly higher accuracy than \texttt{baseline}, despite using similar total computation. This indicates that adjusting Rollout \(N\) helps improve computational efficiency in practice.


\end{document}